\documentclass[conference]{IEEEtran}
\usepackage{times}

\usepackage[numbers]{natbib}
\usepackage{multicol}
\usepackage[bookmarks=true]{hyperref}
\usepackage{graphicx}
\usepackage{subfigure}
\usepackage{todonotes}

\begin{document}

% paper title
\title{A versatile robotic hand with 3D perception, force sensing for autonomous manipulation}

\author{\authorblockN{Nikolaus Correll,
Dylan Kriegman,
Stephen Otto and
James Watson}
\authorblockA{Department of Computer Science, University of Colorado Boulder, Boulder, Colorado 80309--0430\\ Email: ncorrell@colorado.edu}}

\maketitle

\begin{abstract}
We describe a force-controlled robotic gripper with built-in tactile and 3D perception. We also describe a complete autonomous manipulation pipeline consisting of object detection, segmentation, point cloud processing, force-controlled manipulation, and symbolic (re)-planning. The design emphasizes versatility in terms of applications, manufacturability, use of commercial off-the-shelf parts, and open-source software. We validate the design by characterizing force control (achieving up to 32N, controllable in steps of 0.08N), force measurement, and two manipulation demonstrations: assembly of the Siemens gear assembly problem, and a sensor-based stacking task requiring replanning. These demonstrate robust execution of long sequences of sensor-based manipulation tasks, which makes the resulting platform a solid foundation for researchers in task-and-motion planning, educators, and quick prototyping of household, industrial and warehouse automation tasks.  
\end{abstract}

\IEEEpeerreviewmaketitle

\section{Introduction}
\begin{figure}
    \centering
    \includegraphics[width=0.8\columnwidth]{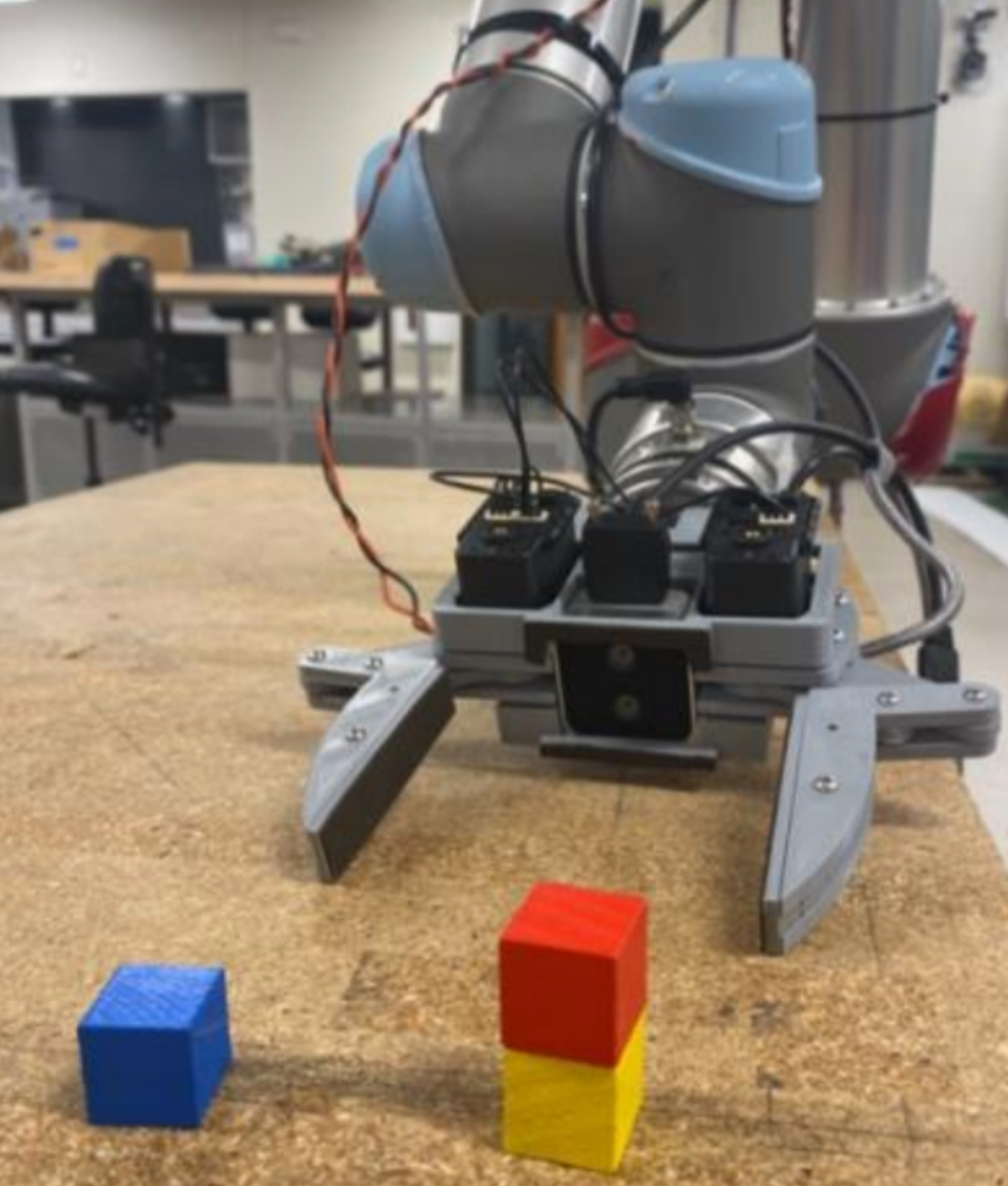}
    \caption{All-in-one manipulation architecture that emphasizes manufacturability, active compliance, robust execution, and versatility. A light-weight software architecture combines object detection, segmentation, point cloud processing, with planning, inference and execution, creating a platform for research in task-and-motion planning for complex manipulation problems.}
    \label{fig:overview}
    \vspace{-2em}
\end{figure}

Autonomous manipulation remains a challenge in digitizing and automating value streams from manufacturing to recycling. These tasks include picking, placing, assembly/disassembly, and packing. They involve a large variety of compliance, for example when tightening a metal nut vs.\ grasping a delicate raspberry. Often, tasks of varying dexterity and compliance need to be combined, requiring a highly versatile hardware and software system. Implementing such a system is difficult. Commercially available grippers have limited capabilities, particularly the absence of torque control, and therefore lack of active compliance \cite{albu2008soft}. Integration with vision systems is not straightforward; wrist-mounted, table-top, and palm-mounted cameras can be occluded at different times during operation. Finally, software frameworks such as ROS take considerable effort to commission and maintain, which is often unnecessary in particular when the goal is to investigate only one aspect of a manipulation pipeline such as vision or high-level reasoning. %\cite{liu2022structdiffusion}.  
In this paper, we present a lightweight manipulation architecture geared at researchers, educators, and hobbyists, that is able to perform a wide variety of multi-step manipulation tasks, while being simple to manufacture and affordable (Figure \ref{fig:overview}).

The ``gold standard'' in manipulation is the human hand, which is able to perform an extensive range of tasks, including tool usage and in-hand manipulation \cite{coulson2021elliott}. Prominent designs in this category are the ``ShadowHand'' that is able to manipulate a Rubik's cube \cite{andrychowicz2020learning}, the RBO hand \cite{bhatt2022surprisingly}, and the Pisa/IIT hand \cite{catalano2014adaptive}. These three examples also span a large spectrum of degrees of freedom from 24 (ShadowHand) to the underactuated Pisa hand which has 19 joints driven by only a single actuator. Benchmarking an end-effector design is a challenging problem and subject to research. It is difficult to disentangle grasp planning from perception \cite{bekiroglu2019benchmarking}, robot arm from end-effector performance \cite{morgan2019benchmarking}, and versatility from specialization, e.g. cloth manipulation \cite{garcia2020benchmarking} or warehouse picking \cite{correll2016analysis}.

One way to evaluate the capability of a hand design is to subject it to a wide variety of household tasks \cite{sun2018robotic} that overlap with upper-limb prosthetic applications. In such a test, underactuated designs such as the Pisa hand \cite{bonilla2018advanced} and a 100-year old claw design that has only a single actuated finger \cite{patel2018manipulation} have outperformed all other designs in \cite{sun2018robotic}. Although the Pisa hand's compliance is advantageous when operating scissors, e.g., a simple mechanical design is intriguing due to its manufacturability, robustness, and simpler control \cite{fearing1986simplified} while still being highly versatile.

Although versatility is not important in conventional automation using highly specialized end-effectors such as during warehouse picking or assembly of the same part, versatility will become critical for recovering from errors or when a task not only requires picking, but precise placement, e.g. during packing. 

%Prosthetic applications also highlight the strong interplay between mechanism, perception and intelligent planning. 

In previous work, we have designed an industrial-grade robotic gripper following these design principles that integrated 3D perception, force control and computation \cite{correll2021systems} to pick up items as small as a washer and as large as a commercial toner cartridge. The system is gentle enough to handle a strawberry, and strong enough to tighten a nut. %\cite{watson2020autonomous}. 
In this paper, we describe a design with much-improved manufacturability by relying on commercial, off-the-shelf motors and cameras, forgoing costly custom integration of computation and cabling. Although sacrificing ruggedness and stand-alone capability, this design can be manufactured at a fraction of the cost (around \$460 in parts including camera), and less than half of the weight (414g), allowing operation on a wide range of robotic arms, while requiring only a 3D printer and off-the-shelf parts such as stand-offs and bearings. 

Based on experience with a wide range of robotic competitions ranging from warehouse picking \cite{correll2016analysis}, household automation \cite{patel2018manipulation}, robotic assembly \cite{von2020robots} and mobile manipulation \cite{roa2021mobile}, we have developed a perception and planning pipeline that combines recent advances in low-cost 3D perception with deep learning for instance-segmentation, and conventional planning. Here, the emphasis of our architecture is to robustly deal with sequencing hundreds of behaviors based on visual and tactile cues. The software stack is released under the MIT License\footnote{\url{https://github.com/correlllab/MAGPIE}}. 

\section{Design}
\subsection{Hardware}
Our design is driven by the requirement to perform assembly tasks involving parts as small as M3 nuts driven by the robotic assembly domain \cite{von2020robots} and household tasks as in \cite{sun2018robotic}. Trading strength and size for manufacturability and cost, we designed the system to pick up 88\% of the items in the YCB benchmark \cite{calli2017yale}, excluding the hammer, the power drill, the pan, and other large objects. With this, we were able to specify the gripper aperture and motor strength, assuming a conservative friction coefficient of $\mu=0.4$ and a safety factor of 1.5.

\begin{figure}
    \centering
    \includegraphics[width=0.5\columnwidth]{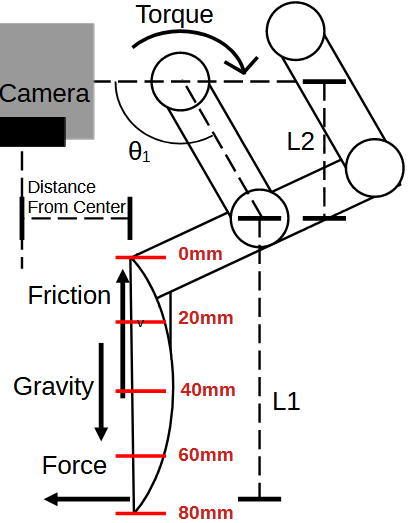}
    \caption{Torque Diagram to Calculate Estimated Force on the Object.}
    \label{fig:torque}
    \vspace{-2em}
\end{figure}

\begin{figure*}
\begin{minipage}{0.49\textwidth}
    \centering
    \includegraphics[width=0.98\columnwidth]{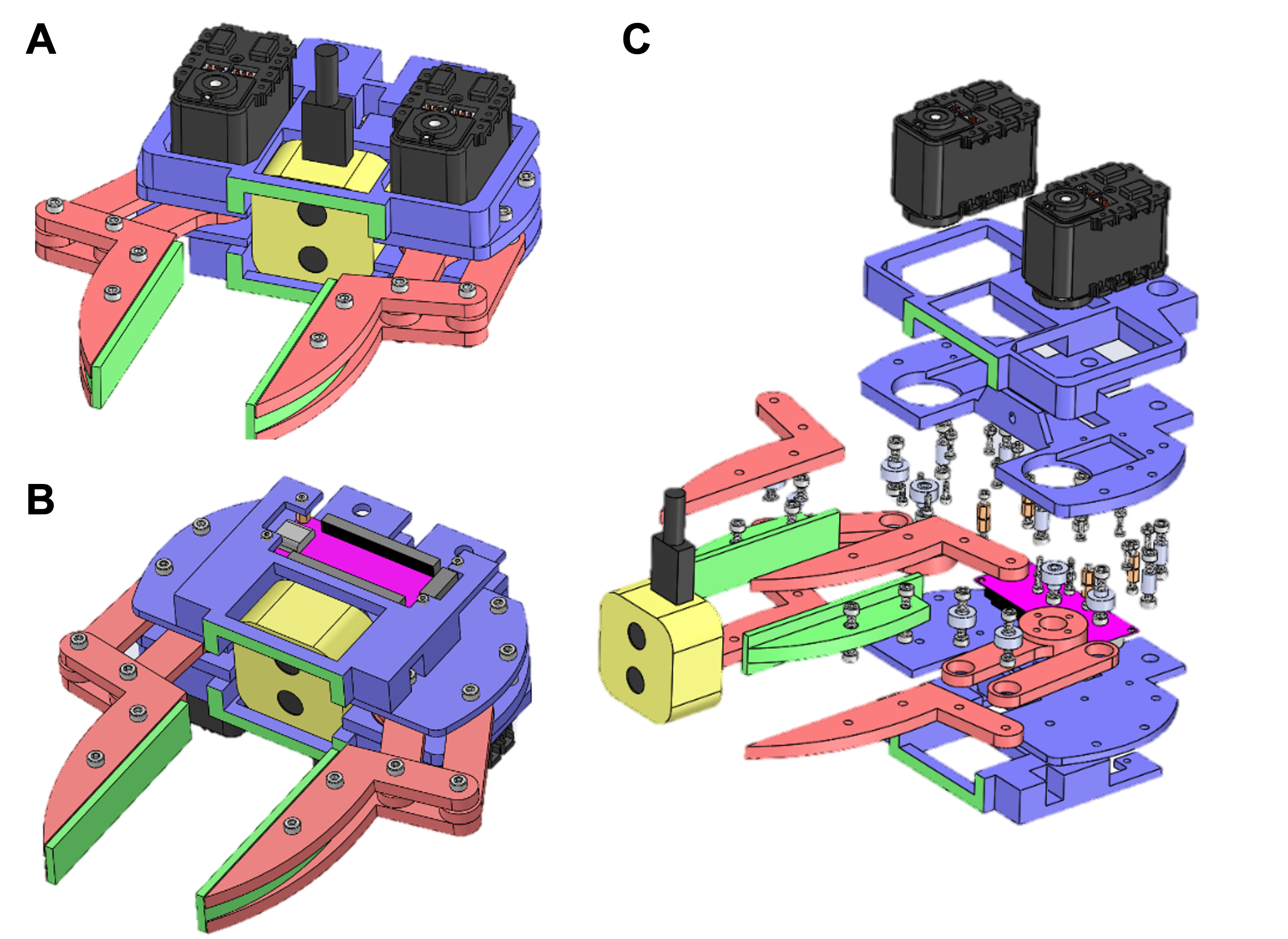}
    \label{fig:my_label}
\end{minipage}
\hfill
\begin{minipage}{0.49\textwidth}
{\fontsize{7}{8}\selectfont
\begin{tabular}{||c l c c||}
 \hline
Bill of Materials \\ [0.5ex] 
 \hline\hline
 ID & Item & Quantity & Cost \\
 \hline\hline
 1 & Top Base(PLA) & 1 & \$ 1.51 \\ 
 \hline
  2 & Top Base Cover(PLA) & 1 & \$ 2.05 \\ 
 \hline
 3 & Bottom Base(PLA) & 1 & \$ 1.17 \\ 
 \hline
  4 & Bottom Base Cover(PLA) & 1 & \$ 1.15 \\ 
 \hline
 5 & Servo Crank(PLA) & 2 & \$ 0.47 \\ 
 \hline
 6 & Servo Coupler(PLA) & 4 & \$ 1.56 \\ 
 \hline
 7 & Servo Rocker(PLA) & 2 & \$ 0.28 \\ 
 \hline
 8 & Finger & 2 & \$ 0.86 \\ 
  \hline
  9 & OpenRB-150 board & 1 & \$ 24.90 \\ 
 \hline
  10 & Intel® RealSense™ D405 & 1 & \$ 272.00 \\ 
 \hline
  11 & AX-12A Servo Motor & 2 & \$ 99.80 \\ 
  \hline
 12 & 3M 6mm standoff & 4 & \$ 9.99 \\ 
  \hline
13 & 3M 8mm standoff & 8 & \$ 9.99 \\ 
  \hline
14 & 2.5M 10mm standoff & 4 & \$ 9.99 \\ 
  \hline
15 & Electrical Wire & 2 & \$ 0.45 \\ 
  \hline
  16 & 3M Nuts and Bolts & 32 & \$ 12.49 \\ 
  \hline
    17 & 2M Nuts and Bolts & 8 & \$ 9.99 \\ 
  \hline
   18 & 5M bearings& 8 & \$ 9.99 \\ 
  \hline
  \hline
   &  & Total & \$458.63 \\ 
    \hline
\end{tabular}
}
\end{minipage}

\caption{Left: CAD Drawings of the gripper from the top (A), bottom (B), and exploded view (C). Each motor independently actuates a finger, providing independent torque-based control. The camera is integrated into the palm. Right: Bill of Materials with approximated cost. PLA with a cost of \$0.05 per $cm^2$. }
\vspace{-1em}
\end{figure*}

The 4-bar linkage design (Figure \ref{fig:torque}) provides a maximal field of view for the palm-mounted camera (Intel RealSense D405), providing a maximum aperture of $w=106.24mm$. All pieces have been 3D printed using PLA. Although PLA printed parts have lesser stiffness than Delrin or ABS sheets, laser cutters are not available to most educators and hobbyists. All links use M5 bearings that are press-fitted into openings in the PLA material and M3 screws as the axis. We found using bearings to greatly reduce friction, while press-fitting reduces play and thereby increases the accuracy of the mechanism. 

%The overall dimensions of the gripper are a maximum width of 211.56mm and a minimum length of of 169.64mm when the robot gripper is fully open and a minimum width of 133.68mm and a maximum length of 193.32mm when the crank is parallel to the gripper. The height of the gripper is 65mm. The mass of the gripper is 0.411 kg. 

%Figure \ref{fig:torque} shows the dimensions that affect the force F_{Force}$ that a finger in a 4-bar linkage can exert, which depend on the opening angle $\theta$ and motor torque $\tau$:
%\begin{equation}
%Weight = \mu*Force
%\end{equation}
%\begin{equation}
%Force = Torque/distance
%\end{equation}
%\begin{equation}
%Force = Torque/(L2*sin(\theta)+L1)
%\end{equation}
%\begin{equation}
%Normal Force = Force*sin(\theta)
%\end{equation}

%In order for a grasped object to stay static, the gravitational force needs to be equal to the friction force. This lateral force translates into an orthogonal force $F_{Friction}=\mu F_{Force}=\tau/(L_2\sin(\theta)+L_1)$ that counteracts gravity and depends on the friction coefficient $\mu$. With $L_1=1.32mm+finger position$ and $L_2=45mm$, assuming $\mu=0.6$ for the friction of rubber on steel, and a safety margin of 1.5x requires a torque of XXXX Nm to hold a weight of XXXX kg. 

The AX-12 servo motors (Robotis, Korea) have a stall-torque $\tau$ of 1.5Nm. The AX-12's digital interface allows setting and measuring the actual motor torque via current control, thereby enabling compliant control and detecting objects within the gripper. Independently controlling the fingers helps to prevent the gripper to get stuck when one of its fingers is impeded, e.g. when picking items from clutter, as well as grasping objects off center \cite{correll2021systems}.

The Intel RealSense camera has an operational range from 7cm to 50cm with a field of view of 87$^o$ × 58$^o$ and a resolution of 1280 × 720. It can detect objects as small as 0.1mm, making it ideal for in-hand perception applications. The camera is mounted in the palm \cite{correll2021systems}, allowing the hand to see an object until contact is made, minimizing distortion and occlusion, when compared to wrist- and table-mounted cameras.  Figure \ref{fig:example} (center), shows a snapshot through the camera, with only minimal occlusion by the fingers.

\subsection{Software} 
Our grasping pipeline follows the outline described in \cite{correll2022introduction}: objects are detected from RGB camera images using pixel-wise segmentation based on Yolo v5\footnote{\url{https://github.com/ultralytics/yolov5}}. Segmentation masks are then used to isolate 3D point cloud information from individual objects. We determine the orientation of the point cloud using principal component analysis (PCA), compute a grasp pose based on the object's bounding box, and use an inverse kinematics solver from Peter Corke's robotic toolbox \cite{rtb} to move the robotic arm to the desired pose. Together, these components create a robust pipeline in which individual elements can be replaced for research or education. All of the behaviors are embedded into Behavior Tree (BT) \cite{colledanchise2018behavior} nodes that allow for robust engineering of multi-state robot controllers. %This allows running the entire controller in a single thread, which increases efficiency and facilitates debugging, while also remaining modular for easy replacement of individual components of the pipeline that are particular to education or research. 

In order to be able to manage long task sequences comprising hundreds of steps, e.g. as required during assembly of complex mechatronics \cite{von2020robots}, our planner automatically generates BTs from a high-level task description of the scene in PDDL 2.1 \cite{fox2003pddl2}, which can be efficiently solved by the AI planning framework FastDownward \cite{helmert2006fast} that we interface via py2PDDL \cite{py2pddl}. Here, the planning problem is populated from the Yolo v5 object recognition pipeline after every step and the plan is recomputed to catch potential execution failures. Here, spatial relationships between objects (e.g. ``on top'') are hard-coded. This framework is illustrated in Figure \ref{fig:example}, showing an example of successful task completion despite errors during execution by continuous replanning.

\section{Evaluation}
\paragraph{Hardware}
In order to test the torque-controlling ability of the servo, we measure the force at the finger as a function of the dimensionless Dynamixel settings in the range of 0--1023. A force gauge (Shimpo FGV-10XY) was locked into a vice to hold it securely for all measurements. The force gauge gives an error of $\pm.2$\%. The force increases linearly ($R^2 = .995, y = 0.0316x - 3.0194$) from 0 to 9N at ``400'' and then increases to 32N following a second-degree polynomial function ($R^2=.999$). In the linear regime, we are able to change force in increments of 0.08N.  

\begin{figure*}
% \vspace{-2em}
\centering
\includegraphics[width=0.9\textwidth]{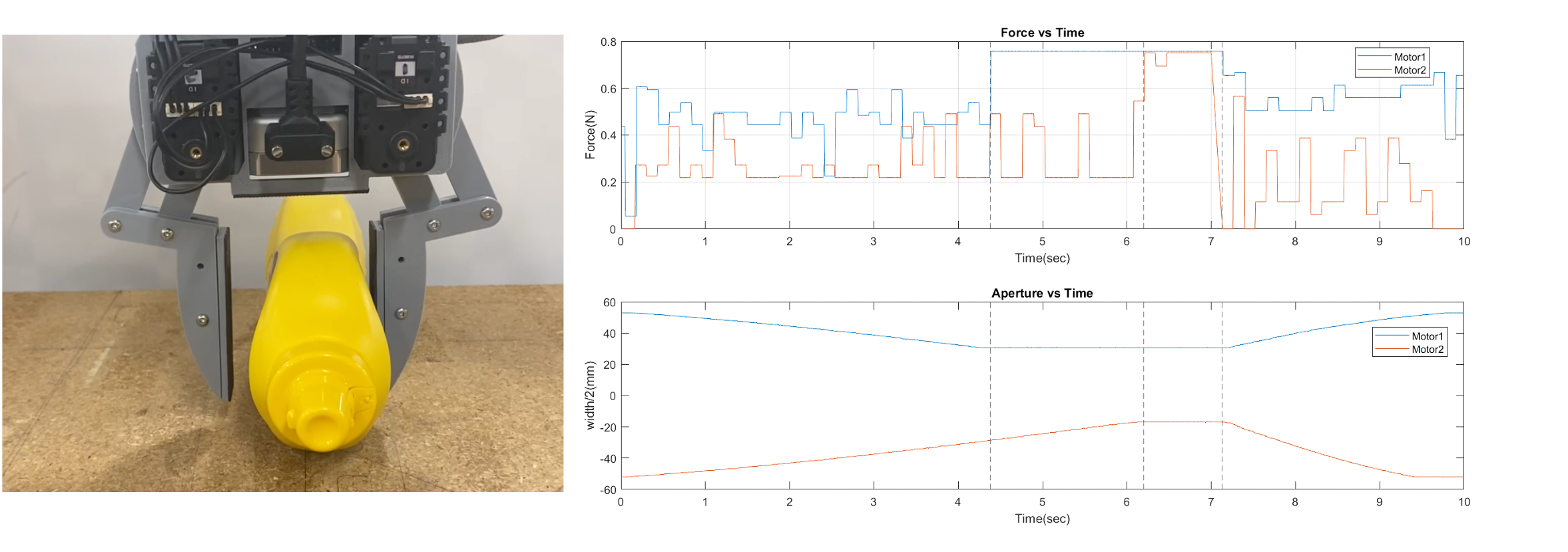}
\vspace{-2em}
\caption{Gripper grasping the mustard container from the YCB dataset from an off-center position (left). Limiting torque during approach prevents the mustard from moving as the right finger makes contact. Dashed vertical lines indicate contact by the right finger, contact with both fingers, and opening (from left to right). }
\label{fig:forcemeasurement}
\end{figure*}

We also demonstrate the gripper's ability to sense force/torque. Figure \ref{fig:forcemeasurement} shows the force profile for the right (`Motor 1') and the left (`Motor 2') finger vs.\ gripper aperture as the gripper closes on the mustard container from the YCB dataset. The container was placed off-center, with the right motor making contact first. Instead of moving the mustard, the finger remains in position until the second finger makes contact. We observe minimal torques while moving the finger, with Motor 1 having more friction than Motor 2 due to variations in manufacturing. 

\begin{figure*}
\centering
    \includegraphics[width=0.9\textwidth]{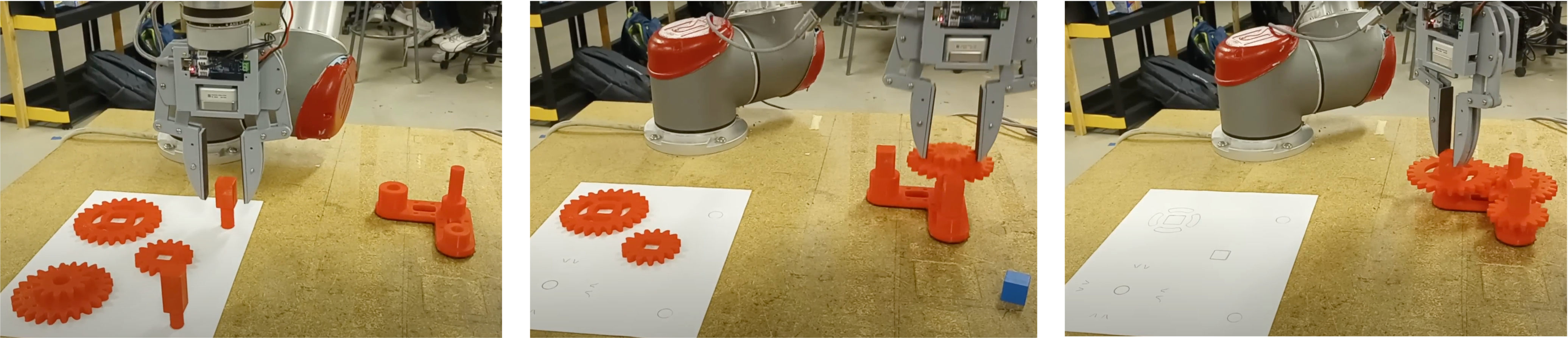}
    \caption{Demonstration of accuracy and precision by reliably assembling the ``Siemens gear assembly problem'' \cite{vecerik2019practical} with sub-millimeter accuracy requirements.}
    \label{fig:gearassembly}
\end{figure*}

To demonstrate the precision and accuracy of the design, we perform a robotic assembly task with tight tolerances ($<0.5mm$). Snapshots of the task are shown in Figure \ref{fig:gearassembly}. Precision and accuracy of grasping the elements from a known position on the kitting mat (modeling the industrial assembly challenge requirements from \cite{von2020robots}) is sufficient to reliably assemble all parts into a functioning mechanical system.  The \emph{open-loop} assembly task succeeded 8 out of 10 trials; with one failure each due to an angular misalignment of the small peg and the large gear becoming jammed on the large peg. These results motivate a sensor-based re-planning approach such as described in \cite{watson2020autonomous}.

\begin{figure*}
    \vspace{-1em}
    \centering
    \includegraphics[width=\textwidth]{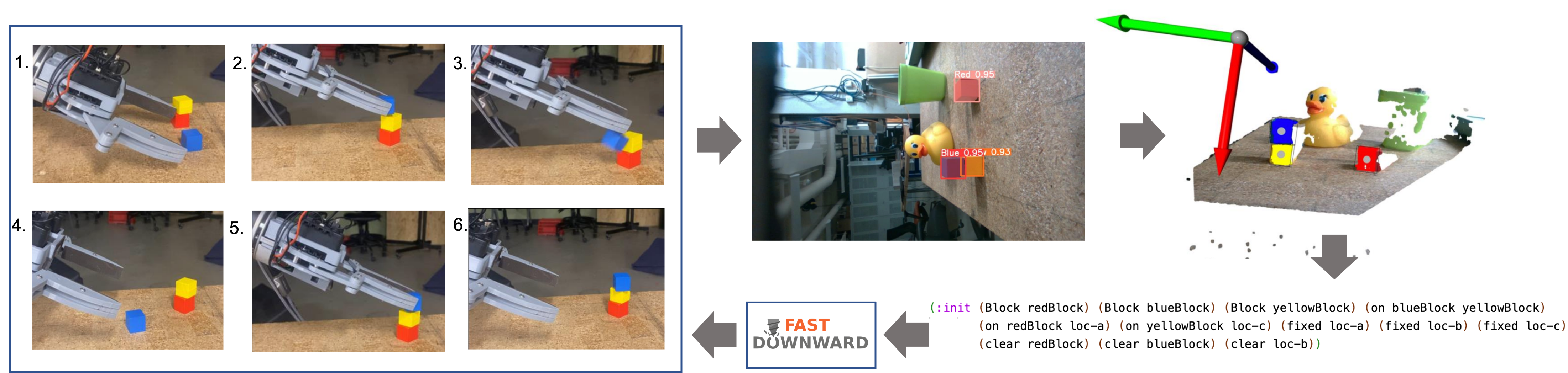}
    \caption{Snapshots from a tower construction task that is solved by continuous replanning. The robot accidentally hits the tower (1) while placing the blue block (2), which fails as the tower has moved (3), the robot re-analyzes the scene (4), and places the blue block (5). Point cloud and image data are used for object identification and segmentation, resulting into a labeled scene that gets parsed into a PDDL 2.1 problem description and solved by FastDownward. The plan is repeated until the problem is solved. 
    }
    \label{fig:example}
    \vspace{-1em}
\end{figure*}

\paragraph{Software}
We evaluated the system using a tower-assembly task using three colored, wooden cubes from \cite{calli2017yale}. Figure \ref{fig:example} shows snapshots of the experiment, a view of the camera image, the resulting labeled 3D objects, and the corresponding PDDL domain. FastDownward automatically generates a plan, which our system translates into a behavior tree to execute the necessary steps. Running this loop once corresponds to the classical sense-plan-act approach. Updating the PDDL domain and replanning accordingly allows for more sophisticated behavior, including reacting to changes in the environment or recovering from errors in execution. We ran a single step, moving one block from one tower to another, 50 times. Of the 42 successful attempts, 30 were completed at the first trial, 6 required two trials, and 3, 2, and 1 experiments required 3, 4, and 6 trials. The other 8 failed in a way that could not be recovered with this simple program, for example when a block was moved out of the field of view of the hand.  

%This system is currently under development with applications in robotic assembly and we will continue contributing to the open-source framework presented here.

\section{Discussion and Conclusion}
The proposed design exceeds the capabilities of \cite{correll2021systems} in terms of perception and in the dimensions of objects that can be picked up. %Choosing more economical servo motors (an order of magnitude less expensive than those used in \cite{correll2021systems}), limits the maximum holding torque; which might be disadvantageous during heavy assembly. 
Yet, 3D-printed PLA plastics cannot reach the level of stiffness and accuracy that a sheet-metal-based design can, which is a disadvantage when grasping very small pieces, such as an M3 screw that requires exact alignment of the fingers.

The current design requires three external cables that need to be routed along a robotic arm: a USB 3.1 connection for the camera, a USB 2.0 connection for the motor control board, and a power cable that can provide up to 3A at 12V (36W). All of these could be met by a single USB-C PD connection, in particular when 5V to 12V step-up circuits with sufficient power are available.  

The emphasis on the software pipeline is to demonstrate modularity (via the BT framework and STRIPS planner), rather than performance on specific subtasks. Indeed, components such as object detection could be easily tuned to achieve 100\% one-shot success rate in the tower assembly task. Failure has been instructive as it will be pervasive in autonomous settings, and remain a problem if it occurs with a non-zero probability. Here, we are particularly interested in reasoning under uncertainty and generating high-level plans for recovery from error using semantic domain knowledge. For example, when an object moves outside of the robot's field of view, the robot could search for it and move objects out of the way if it has to. We plan to further explore this in the future by interfacing PDDL descriptions with large language models to (1) generate problem descriptions from image/3D data, (2) generate PDDL descriptions from natural language, and (3) use domain knowledge to increase the robot's reasoning abilities. Finally, we are interested in building upon this design to perform better on heavy items by adding a thumb, operating tools such as a power-drill by adding an additional degree of freedom to one finger, and fast picking by adding a suction cup.

\section*{Acknowledgements}
This work has been supported by a grant by the National Science Foundation ``USDA-NIFA NRI INT: Autonomous Restoration and Revegetation of Degraded Ecosystems''.
%An important safety consideration in industrial grippers is the ability to hang on to an object in case of power failure. This is not the case in this design, as the servos require power to hold torque. 

%With its ability to reliably execute complex manipulation tasks that are specified using symbolic language, our pipeline is ideally suited to validate research involving the emerging area of large language models, as well as a basis for improving grasping and manipulation hardware itself. Here, developing appropriate benchmarks and specifying requirements on autonomous robotic systems  remains an open challenge.

%\todo[inline]{I don't think the idea that the system can operate on semantic instructions from LLM is well supported, and this is only briefly mentioned at the very beginning and the very end. Consider dropping it?}

%\section{Conclusion}

% DISCUSS WHY WE CAN'T 

% \newpage
% \bibliographystyle{plainnat}
\bibliographystyle{abbrvnat}
% \vspace{-1em}
\bibliography{references}
\end{document}